# Decoupled Federated Learning on Long-Tailed and Non-IID data with Feature Statistics

Zhuoxin Chen
*Beijing University of Posts and Telecommunications*
Beijing, China
chenzhuoxin@bupt.edu.cn

Zhenyu Wu*
*Beijing University of Posts and Telecommunications*
Beijing, China
shower0512@bupt.edu.cn

Yang Ji
*Beijing University of Posts and Telecommunications*
Beijing, China
jiyang@bupt.edu.cn

*Abstract*—Federated learning is designed to enhance data security and privacy, but faces challenges when dealing with heterogeneous data in long-tailed and non-IID distributions. This paper explores an overlooked scenario where tail classes are sparsely distributed over a few clients, causing the models trained with these classes to have a lower probability of being selected during client aggregation, leading to slower convergence rates and poorer model performance. To address this issue, we propose a two-stage Decoupled Federated learning framework using Feature Statistics (DFL-FS). In the first stage, the server estimates the client's class coverage distributions through masked local feature statistics clustering to select models for aggregation to accelerate convergence and enhance feature learning without privacy leakage. In the second stage, DFL-FS employs federated feature regeneration based on global feature statistics and utilizes resampling and weighted covariance to calibrate the global classifier to enhance the model's adaptability to long-tailed data distributions. We conducted experiments on CIFAR10-LT and CIFAR100-LT datasets with various long-tailed rates. The results demonstrate that our method outperforms state-of-the-art methods in both accuracy and convergence rate.

## 1. Introduction

The rapid evolution of Internet technology has underscored the increasing significance of data, while concurrently amplifying concerns regarding data security and privacy. To tackle these challenges, federated learning (FL) technology has emerged and found extensive application across various domains, including corporate entities and diverse device ecosystems [1][2]. For instance, financial institutions leverage federated learning to infer consumer purchasing behavior for tailored financial services, while medical establishments employ private computing for more sophisticated medical diagnostic systems.

However, a significant challenge faced in practical federated learning applications is data heterogeneity. The global dataset often displays long-tailed distributions [6], while the data among clients is commonly non-independent and identically distributed (non-IID), especially in terms of label distribution skew. The difference in distribution between clients leads to a scenario that is often overlooked in many studies: for tail classes, not only is the overall data volume limited, but the number of clients with data for that tail class is also relatively small. For example, certain medical conditions might have limited occurrences, primarily diagnosed by specialized hospitals. An illustrative case is evident in Figure 1, showcasing the natural distribution from the EMNIST-letters dataset. The letter "O" represents the head class, while "K" signifies the tail class. When randomly distributing these data into blocks of 500 samples across 20 clients, it becomes evident that not only is the overall amount of tail class data limited, but also the number of clients holding such data is minimal. As a result, when clients are randomly selected in each round, those with the tail class data have a lower probability of being chosen. This tendency leads the aggregated global model to consistently favor the head class, resulting in significant information loss related to the tail class. This trend results in slower model convergence during training, increased training costs, and ultimately leads to lower accuracy and poorer quality in the final model.

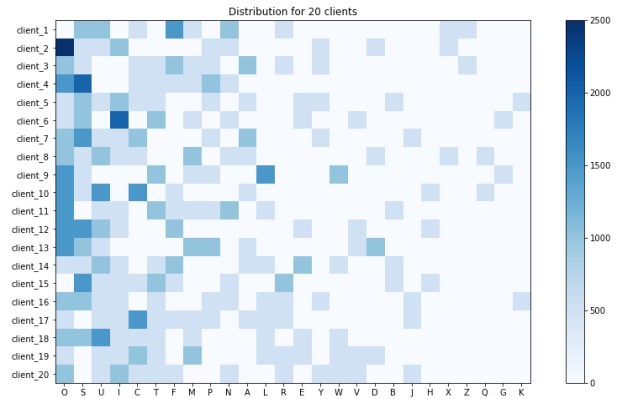

**Fig. 1.** Heat map for client data distribution under long-tail distribution and non-IID in EMNIST-letters dataset. Not only the overall data volume of tail classes is small, the number of clients with data for tail class is also relatively small.

Previous studies [14][15] solve long-tailed data in federated learning by estimating the global imbalance ratio to design the loss function, but the establishment of their methods requires ensuring that the client distribution is IID. Although other studies on the non-IID problem [3][4][5] consider the distribution differences between clients, they assume that the global data distribution is balanced and ignore the reasons why long-tail distribution may degrade the model. Recently, several studies have highlighted classifier degradation as a pivotal factor contributing to the challenges posed by global long-tailed distribution and non-IID data in federated learning. In the study by CCVR[16], an exploration into non-IID effects revealed a disparity in similarity among different clients within deeper layers. They propose leveraging classifier calibration to mitigate bias by generating virtual features using a Gaussian distribution. Building upon this, CReFF[17] delved deeper into the issue of long-tail distribution. They discovered that retraining individual layers of the FedAVG[1] network with balanced data significantly enhanced performance, emphasizing the substantial gains derived from classifier retraining. However, due to privacy constraints, accessing the original balanced data from each client is restricted. To tackle this, CReFF introduced a viable two-stage learning framework, which utilized gradients uploaded by clients to synthesize feature sample data on the server side, facilitating classifier retraining and subsequent updates in global training rounds. However, in cases where tail classes are sparsely distributed among clients, randomly selecting clients to upload gradients may lead to incomplete information transmission for these classes. This can result in missing categories in regenerated federated features, impacting classifier learning and potentially introducing bias toward head classes. Besides, CReFF have to pass class distribution information, which may lead to client-side security attacks and privacy leaks.

---
* Zhenyu Wu is the corresponding author.

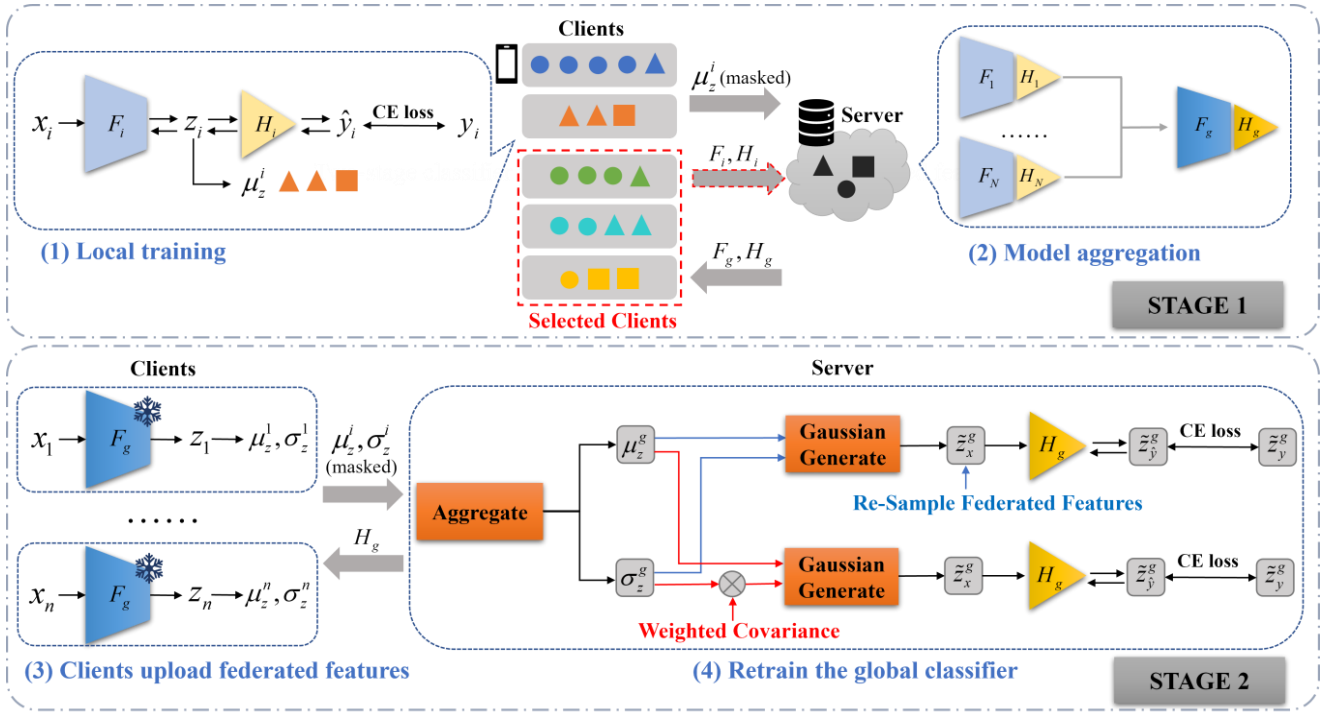

**Fig. 2.** Proposed framework of DFL-FS. We decouple the model into a feature extractor $F$ and a classifier $H$ within a two-stage training framework. In the first stage (1) the clients train the local model and calculate feature statistics $\mu_z$. (2) The server then estimates the client's coverage distribution through clustering masked feature statistics and filters class-balanced clients for model aggregation. In the second stage, (3) the clients use the frozen global feature extractor to calculate local feature statistics, and upload them to the server, (4) the server then calculates the global feature statistics to regenerate samples from Gaussian distribution by resampling or weighted covariance to retrain the classifier.

Motivated by the above findings, we propose a Decoupled Federated learning framework with Feature Statistics (DFL-FS) to solve long-tail distribution and non-IID problems. Specifically, we decouple the model into a feature extractor and classifier, focusing on feature learning in the first stage, while in the second stage, the local client generates federated features from a frozen feature extractor and the server regenerates samples to retrain the global classifier. In the first stage, to solve the problem of biased client distribution caused by random sampling, we proposed a client selection strategy based on feature statistics clustering. The unknown classes on the client are used as feature masks to further enhance privacy protection. In the second stage, we calculate the feature statistics of each global class and then use Gaussian distribution to generate feature samples for retraining the classifier. We found that even the classifier trained with a balanced number of samples was still biased, so we proposed two strategies: weighted covariance and resample features to enhance the model's focus on tail classes.

Our main contributions are summarized below:

- We introduce DFL-FS, addressing long-tail distribution and non-IID challenges using a two-stage training approach with feature statistics.
- We propose a client selection strategy MFSC based on masked local feature statistics clustering to select suitable clients before aggregating the model to improve convergence.
- We propose weighted covariance and resampling federated features based on global feature statistics to enhance classifier retraining.
- We verified the effectiveness of the DFL-FS method on the public image data sets CIFAR10-LT and CIFAR100-LT and achieved SOTA.

## 2. RELATED WORK

### 2.1. Federated learning with non-IID data

As a typical non-IID scenario, the problem of the label distribution skew is widely studied in FL. FedAVG[1] was first proposed to solve this problem, and some other methods have been improved based on it to control the model update amplitude by adding regular terms, such as FedProx[3], FedNova[4], and MOON[5]. CCVR [16] proposed a post-hoc method to improve the effect through classifier calibration. However, these methods assume that the global data is balanced and randomly select clients before doing federated aggregation. In practice, data often follows a long-tailed distribution and the tail class tends to be sparsely distributed across a few clients. Randomly selecting clients for aggregation in each round may overlook these tail classes, leading to slow model convergence and deviations from the intended convergence direction[18].

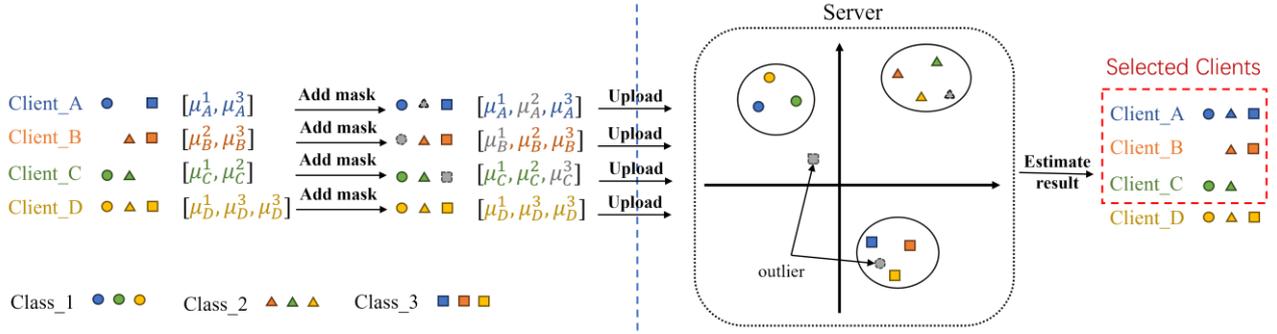

**Fig. 3.** Masked Feature Statistics Clustering (MFSC) on the server side to estimate clients' coverage distribution. The client adds mask for non-existent categories to prevent privacy leakage. The server clusters all feature points. If a certain category appears outlier, it is judged that the client does not have that category, thereby estimating the client's category distribution.

**2.2. Federated learning with long-tailed learning**

Previous research mainly focuses on the long-tailed problem under local centralized learning, which can be divided into three main categories[6]: (1) class re-balancing[7], (2) data augmentation[9][10], (3) module improvement[11][12]. These methods can work under centralized training conditions but may not be feasible in federated scenarios due to privacy constraints, like sampling and data augment require raw data access. Some methods try to solve the long-tail problem in federated learning from the perspective of adjusting the loss function. Fed-Focal[14] introduces focal loss in the model training process, but since the clients participating in each round change, it is difficult to know the data distribution in advance. Ratio-loss[15] proposes a method of estimating data distribution using gradient magnitude but requires the auxiliary data set, which is often limited or incompletely collected. CReFF[17] first proposes a two-stage training framework, by retraining the classifier by generating balanced feature samples using gradients on the server side. However, in the first stage of feature learning, they use random client selection, which easily causes the tail class to be selected with a lower probability, resulting in the loss of its corresponding gradient, and making the retrained classifier biased towards the head class.

## 3. METHODOLOGY

### 3.1. Problem setting and proposed framework

We consider a cross-device federated learning scenario where the participants consist of $N$ clients and one server. Each client has a local dataset with no duplicates, following a non-IID distribution $P_i(x, y), x \in D_i, y \in C, i = 1, \ldots, K$, where C denotes the total class number, and the whole data $D = \bigcup D_i$ is long-tailed. We decouple the model into a feature extractor ($F$) and a classifier ($H$). Specifically, the whole parameters are denoted as $w = [w_f; w_h]$. We denote the feature statistics as the mean $\mu$ and covariance $\sigma$ of features $z$ extracted by the feature extractor for each class.

The framework of our proposed method DFL-FS is shown in Fig. 2, follows a two-stage framework. In the first stage, individual clients conduct local training on private data, simultaneously computing and uploading masked feature means for each class to the server. The server employs local feature statistics($\mu_i$) clustering to select suitable models for aggregation, which are then sent back to clients for iterative training until the convergence of the global model. The second stage involves a single client-server interaction for classifier retraining. Clients use the pre-trained frozen global model to extract and upload masked feature statistics($\mu_i, \sigma_i$). The server estimates the true client's coverage distribution, computes the global class-wise mean($\mu_g$) and covariance($\sigma_g$) of features, and leverages a Gaussian distribution to regenerate sample features for retraining the global classifier. To prioritize tail class training, we employ weighted covariance adjustment and resampling before and after Gaussian distribution. The retrained classifier is subsequently sent back to each client, together with the feature extractor from the initial stage to complete the model.

### 3.2. Client selection based on local feature statistics

In the context of long-tailed global data distribution, where both the overall data volume and the number of clients in the tail class are relatively small, it is essential to address the issue of imbalanced model aggregation during federated learning. Previous work [18] suggests that a suitable client sampling strategy should ensure unbiasedness, meaning that the expected aggregation value of a subset of clients should be equal to the aggregate value when all clients are available.

To achieve this, we propose a client selection strategy based on masked feature statistics clustering (MFSC), as illustrated in Figure 3. In addition to the local model ($F_i, H_i$), clients upload local feature statistics ($\mu_i$) to the server. To protect privacy, we employ feature masking by assigning random numbers to feature means corresponding to categories that do not exist on the local client (as shown in the gray graphic in Figure 3). This prevents the leakage of the client's class coverage distribution information and mitigates potential network attacks from the client side.

The server can obtain the feature statistics of all local classes uploaded by the client and then perform clustering. As shown in Fig. 3, if there is an outlier after clustering for a certain class, it is judged that the client holding the feature point does not have this class. If the random feature points happen to be distributed within the clustering range, it will also be judged that the category exists in the client. Then the server prioritizes client selection based on the distribution of class coverage, starting from the smallest tail class to the head class. This approach aims to attain comprehensive and balanced class representation across clients in model aggregation. Although the number of clients participating in each round is limited, the feature statistics will not change drastically, so the server can maintain and update the feature statistics last uploaded by each client for estimation and filtering.

**3.3. Classifier retraining based on global feature statistics**

In consideration of privacy protection, local balancing strategies are often no longer applicable, such as re-sampling have to access the original data, and re-weighting have to know the global data distribution in advance and inform all clients, which seriously violates the original intention of federated learning. In this case, it is necessary to additionally select shareable information as the basis for retraining the classifier. CReFF uploads gradients from each client and regenerates features on the server to retrain the classifier. However, the client's class distribution information also needs to be uploaded, which may cause potential information leakage. We follow the idea of using feature statistics to regenerate federated features and conduct an experiment to see whether this method is still applicable to the problem of long-tailed distribution.

We first use the MNIST dataset to do a simple toy experiment and set the model feature output to 2 dimensions to view the feature distribution and draw the features generated according to the global feature statistics. We implement a standard federated training with 100 clients and set the imbalance factor as 0.02, with the non-IID factor as 0.5. Then we extract features from each local client's train set to calculate global mean $\mu_g$ and covariance $\sigma_g$ and regenerate federated features from Gaussian distribution as (1)(2), where $N_i$ is the number of samples in client $i$ and $N_g = \sum N_i$. We use the regenerated federated features to retrain the global classifier and test it on the test set and draw the features of the test set and its classification results, as shown in Fig. 4(a). Note that we generated a consistent number of samples for each class during retraining, but it can be seen from the feature distribution that the feature space of the tail class is smaller than that of other classes, and there exists a large amount of overlap.

$$\mu_g = \sum_{i=1}^{K} \frac{N_i}{N_g} \mu_i \quad (1)$$

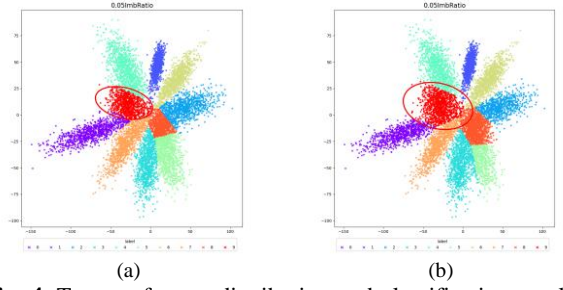

**Fig. 4.** Test set feature distribution and classification results. The part circled in red is the tail class, which has a small discrimination space and overlaps with the head class in (a). The results of retraining using samples generated by our proposed strategy effectively increase the discriminative space of tail classes in (b).

$$\sigma_g = \sum_{i=1}^{K} \frac{N_i - 1}{N_g - 1} \sigma_i + \sum_{i=1}^{K} \frac{N_i}{N_g - 1} \mu_i \mu_i^T - \frac{N_g}{N_g - 1} \mu_g \mu_g^T \quad (2)$$

We wonder what will happen if we make the model pay more attention to the tail class. So, we continued to experiment on this basis and adopted two different strategies:

- **Re-Sample Features (RS):** Artificially constructing an imbalanced distribution for federated features, making the number of tail classes more and head classes less. Thus, the model is better equipped to distinguish tail classes from other classes when feature overlap exists. We sort all classes to a sequence number in advance, and the sequence index ranges from $0...,C$, i.e. the smaller the number of classes, the larger the index. For simplicity, we design a linear formula for regenerated sample number $n_c$ as (3), where base sample size $n_b$ and growth rate $n_k$ are hyperparameters, and $c$ denotes the current class index.

$$n_c = n_b + n_k \cdot c \quad (3)$$

- **Weighted Covariance (WC):** Apply weights to the covariance corresponding to each class according to the global class distribution, the head class applies a small weight, and the tail class applies a large weight. Because the covariance determines the range of feature distribution, increasing the covariance to the tail class makes it occupy more space when generating features through Gaussian distribution. We design a linear formula for weight $w_c$ as (4), where base weight size $w_b$ and growth rate $w_k$ are hyperparameters, and $c$ denotes the current class index.

$$w_c = w_b + w_k \cdot c \quad (4)$$

It's surprising to find that both two strategies are simple but effective. As Fig. 4(b) shows when we conduct a Re-Sample of Features with $n_b = 500$ and $n_k = 150$, the feature space of the tail class is enlarged and closer to the real distribution. We also experiment on CIFAR10-LT to verify the validity of our method. The experimental details are consistent with Section 4.3. and we record accuracy for all 10 classes in Fig. 6, which confirms the correctness of our conjecture.

TABLE I. TEST ACCURACY (%) OF DIFFERENT METHODS ON CIFAR10-LT AND CIFAR100-LT

|  | CIFAR10-LT | | | CIFAR100-LT | | |
| --- | --- | --- | --- | --- | --- | --- |
|  | IF=10 | IF=50 | IF=100 | IF=10 | IF=50 | IF=100 |
| FedAVG[1] | 0.6233 | 0.5162 | 0.4561 | 0.3408 | 0.2507 | 0.2177 |
| Focal-loss[14] | 0.6049 | 0.4988 | 0.4374 | 0.3328 | 0.2385 | 0.1906 |
| Ratio-loss[15] | 0.6509 | 0.5628 | 0.4936 | 0.3516 | 0.2548 | 0.2220 |
| CCVR[16] | 0.7127 | 0.6291 | 0.5760 | 0.3721 | 0.2979 | 0.2706 |
| CReFF[17] | 0.7237 | 0.6347 | 0.5831 | 0.2510 | 0.1679 | 0.1613 |
| DFL-FS (WC) | **0.7622** | **0.6804** | **0.6196** | **0.4176** | **0.3303** | **0.2913** |
| DFL-FS (RS) | **0.7654** | **0.6838** | **0.6245** | **0.4223** | **0.3358** | **0.3127** |

## 4. EXPERIMENTS

### 4.1. Experimental Setup

*1) Dataset:* We mainly experiment on the CIFAR10-LT and CIFAR100-LT datasets. We split the data into blocks, each block has 50 samples. For the long-tailed setting, we first divide the dataset distribution into imbalance factor (IF)={10, 50, 100}. For the non-IID setting, we set $\alpha$ as 0.5 and divide data into 100 clients.

*2) Model:* We use ResNet-8 as the backbone model for the feature extractor and a single linear layer as the classifier. The first stage training step selects 20 clients for aggregation and lasts 200 rounds. Each client trains locally for 10 epochs using SGD with a learning rate of 0.1 and updates parameters with cross-entropy loss. For the second stage training step, we initialize a new classifier model with the same structure as the ResNet-8 classifier and train it using federated features for 100 epochs with SGD and a learning rate of 0.01. The trained classifier parameters will then replace the parameters of the original global model.

### 4.2. Main results

TABEL II. ABLATION STUDY ON MFSC

(a) results on CIFAR10-LT

|  | CIFAR10-LT | | |
| --- | --- | --- | --- |
|  | IF=10 | IF=50 | IF=100 |
| DFL-FS (WC) w.o. MFSC | 0.7516 | 0.6635 | 0.6032 |
| DFL-FS (RS) w.o. MFSC | 0.7537 | 0.6689 | 0.6158 |
| DFL-FS (WC) w.t. MFSC | 0.7622 | 0.6804 | 0.6196 |
| DFL-FS (RS) w.t. MFSC | 0.7654 | 0.6838 | 0.6245 |

(b) results on CIFAR100-LT

|  | CIFAR100-LT | | |
| --- | --- | --- | --- |
|  | IF=10 | IF=50 | IF=100 |
| DFL-FS (WC) w.o. MFSC | 0.4024 | 0.3181 | 0.2783 |
| DFL-FS (RS) w.o.MFSC | 0.4126 | 0.3209 | 0.2894 |
| DFL-FS (WC) w.t. MFSC | 0.4176 | 0.3303 | 0.2913 |
| DFL-FS (RS) w.t. MFSC | 0.4223 | 0.3358 | 0.3127 |

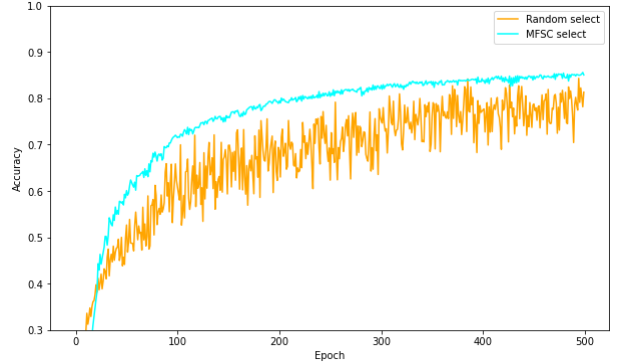

**Fig. 5.** Test set accuracy curve with training epoch. As the number of training rounds increases, the method of randomly selecting clients oscillates seriously, while our MFSC converges quickly and has high accuracy.

We compare our method with FedAVG, Focal-loss, Ratio-loss, CCVR, and CReFF, the results are shown in Table I. From the experimental results, we can see that under different long-tailed ratios, both our two methods achieve SOTA. Since Focal-loss and Ratio-loss do not consider the client distribution on non-IID, the global category weight estimated in the loss calculation is not applicable to each local client, and the result is comparable as FedAVG. Our methods improve by an average of 5% compared to CReFF, and CCVR performs better than CReFF in CIFAR100-LT, a more challenging data set, which demonstrates the advantage of using feature statistics instead of gradients as basis for regenerating samples. Compared with CCVR, we achieved nearly 4% improvement, because we implemented a more reasonable client selection method and adopted a balanced strategy in the second stage.

### 4.3. Module Study

*1) Evaluation of client selection strategy:* We took out the first stage of the framework for a separate experiment, compared the proposed selection strategy based on feature statistics clustering with the strategy of randomly selecting clients, and plotted the test accuracy on each round of test sets. The experimental results are shown in Fig. 5. The results show that our method can make the feature learning training phase converge faster, which means less training time is needed to achieve the same effect, and training costs are reduced. In addition, compared with random sampling, the client-side selection strategy has a higher accuracy when it reaches convergence. Besides, we conducted ablation experiments on the framework as in Table II, and compared whether to adopt a one-stage client selection strategy on the two data sets and found that both resulted in a 1% improvement, proving that our strategy is effective.

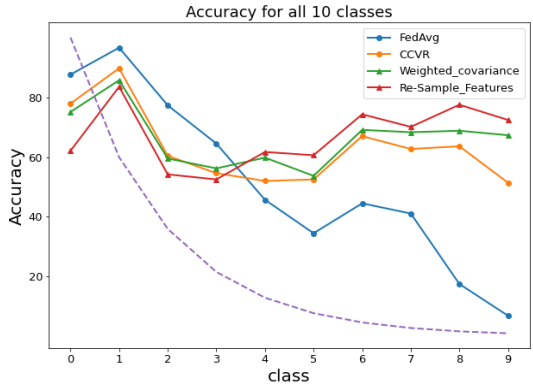

**Fig. 6.** Test accuracy for each class

  *2) Evaluation on classifier retraining:* We experimented with the two federated features optimization strategies proposed in the second stage, compared FedAVG and CCVR, and tested them on each class. For RS, we set $n_b = 500$ and $n_k = 150$, and for WC, we set $w_b = 0.5$ and $w_k = 0.1$. The results are shown in Fig. 6. It can be seen that compared with FedAVG, CCVR and our strategy can effectively improve the test accuracy of the tail class due to the two-stage retraining. Compared with CCVR, both covariance weighting and resampling features can once again improve the recognition accuracy of tail classes, and the improvement brought by them is far greater than the loss of recognition performance of head classes.

## 5. CONCLUSION

In our research, we delve into the impact of long-tailed and non-IID data on federated learning, particularly in scenarios where tail classes are sparsely distributed among clients. This setup affects model convergence as tail classes participate in aggregation with lower frequency, influencing the model's convergence rate and overall effectiveness. To address this, we propose MFSC to filter clients for aggregation rounds and introduce RS and WC techniques aimed at calibrating the classifier to better enhance the model's performance on tail classes. Both strategies are implemented in the DFL-FS framework. Experiment results show that our strategy has a high accuracy and convergence rate while protecting privacy.